\documentclass[conference,final]{IEEEtran}
\IEEEoverridecommandlockouts
\usepackage{cite}
\usepackage{amsmath,amssymb,amsfonts}
\usepackage{algorithmic}
\usepackage{graphicx}
\usepackage{textcomp}
\usepackage{xcolor}

\usepackage{comment}
\usepackage{hyperref}
\hypersetup{hidelinks}
\usepackage{arydshln}
\usepackage{multirow}

\ifCLASSOPTIONcompsoc
\usepackage[caption=false,font=normalsize,labelfont=sf,textfont=sf]{subfig}
\else
\usepackage[caption=false,font=footnotesize]{subfig}
\fi

\usepackage{amsthm}
\theoremstyle{definition}
\newtheorem{definition}{Definition}[section]

\DeclareMathOperator*{\argmax}{arg\,max}

\usepackage[mathlines,switch]{lineno}

\makeatletter

\let\old@ps@IEEEtitlepagestyle\ps@IEEEtitlepagestyle
\def\confheader#1{%
    \def\ps@IEEEtitlepagestyle{%
        \old@ps@IEEEtitlepagestyle%
        \def\@oddhead{\strut\hfill#1\hfill\strut}%
        \def\@evenhead{\strut\hfill#1\hfill\strut}%
    }%
    \ps@headings%
}
\makeatother

\confheader{%
        \parbox{20cm}{\textbf{\textit{2022 9th Swiss Conference on Data Science (SDS) $|$ June 2022 $|$ \textcopyright2022 IEEE $|$ DOI: \href{https://doi.org/10.1109/SDS54800.2022.00011}{10.1109/SDS54800.2022.00011}}}}
}

\def\BibTeX{{\rm B\kern-.05em{\sc i\kern-.025em b}\kern-.08em
    T\kern-.1667em\lower.7ex\hbox{E}\kern-.125emX}}
\begin{document}

\title{Group Fairness in Prediction-Based Decision Making: From Moral Assessment to Implementation}

\author{\IEEEauthorblockN{\href{https://orcid.org/0000-0003-2019-4829}{Joachim Baumann}}
\IEEEauthorblockA{
\textit{University of Zurich}\\
\textit{Zurich University of Applied Sciences}\\
Zurich, Switzerland \\
\href{mailto:baumann@ifi.uzh.ch}{baumann@ifi.uzh.ch}}
\and
\IEEEauthorblockN{\href{https://orcid.org/0000-0002-6683-4150}{Christoph Heitz}}
\IEEEauthorblockA{
\textit{Zurich University of Applied Sciences}\\
Winterthur, Switzerland \\
\href{mailto:christoph.heitz@zhaw.ch}{christoph.heitz@zhaw.ch}}
}

\maketitle

\begin{abstract}

Ensuring fairness of prediction-based decision making is based on statistical group fairness criteria.
Which one of these criteria is the morally most appropriate one depends on the context, and its choice requires an ethical analysis.
In this paper, we present a step-by-step procedure integrating three elements: (a) a framework for the moral assessment of what fairness means in a given context, based on the recently proposed general principle of ``Fair equality of chances'' (FEC) (b) a mapping of the assessment's results to established statistical group fairness criteria, and (c) a method for integrating the thus-defined fairness into optimal decision making.
As a second contribution, we show new applications of the FEC principle and show that, with this extension, the FEC framework covers all types of group fairness criteria: independence, separation, and sufficiency.
Third, we introduce an extended version of the FEC principle, which additionally allows accounting for morally irrelevant elements of the fairness assessment and links to well-known relaxations of the fairness criteria.
This paper presents a framework to develop fair decision systems in a conceptually sound way, combining the moral and the computational elements of fair prediction-based decision-making in an integrated approach.%
\footnote{Data and code to reproduce our results are available at \href{https://github.com/joebaumann/fair-prediction-based-decision-making}{https://github.com/joebaumann/fair-prediction-based-decision-making}.}

\end{abstract}


\section{Introduction}

Recent advances in predictive modeling have led to more and more practitioners using supervised machine learning (ML) techniques to automate decisions in many domains such as allocation of loans, pretrial release decisions, university admissions, or hiring.
However, a prediction does not immediately lead to a decision, and people tend to have different perspectives regarding what constitutes \textit{good} decisions in a particular context.
On the one hand, for a decision maker, it is rational to make decisions that maximize their expected utility.
On the other hand, from a societal perspective, the goal of prediction-based decision systems might be different, e.g., that the decisions do not produce social injustice for the decision subjects (i.e., the individuals affected by the decision).
Especially when prediction-based systems are used to make consequential decisions, issues of fairness might arise.
However, there are many possible sources of unfairness, making it more challenging to ensure fairness for any decision making system~\cite{mitchell2021algorithmic}.
As a result, the field of \textit{algorithmic fairness} has received much interest recently, thereby striving to correct algorithmic biases, which represent systematic and repeatable errors in algorithmic solutions.

Many statistical fairness criteria have been proposed to assess whether decisions based on a predictive model's outputs are systematically biased against certain protected groups such as gender, race, or religion.
Most of these so-called \textit{group fairness criteria} fall into one of three categories: independence, separation, and sufficiency~\cite{barocas-hardt-narayanan}.
In most cases, these criteria are mathematically incompatible~\cite{DBLP:journals/corr/KleinbergMR16,Chouldechova2017}.
Which of the many group fairness criteria is most appropriate depends on the specific decision making context at hand~\cite{Binns2018lessons,Kearns2019EthicalAlgorithm,mitchell2021algorithmic}.
The philosophical literature has been concerned with this choice of a specific definition of fairness (mainly based on egalitarianism principles of distributive justice), but there is no general guideline encompassing all well-known group fairness criteria.
\cite{Loi2019,loi_fair_2021} introduce a fairness principle -- called \textit{Fair Equality of Chances} (FEC) -- to morally justify the choice of any specific group fairness criterion for a particular decision context.
However, the application of the FEC principle is restricted to just two fairness criteria (separation and sufficiency%
\footnote{
In this paper, we correct this mistake and show that the FEC principle also allows for the justification of other group fairness criteria.
}) whereas many more group fairness criteria are established in the fair ML community~\cite{verma2018fairness,narayanan2018translation,castelnovo2021zoo}.
Complicating matters even further, the developers of prediction-based decision making systems are usually not trained to solve ethical problems.
Thus, it is unclear to them how to derive a morally appropriate definition of fairness and how to translate it into a technical solution.

Different techniques have been suggested to implement fair prediction-based decision making systems~\cite{friedler2019comparative}.%
\footnote{
These techniques generally fall under three categories that differ in \textit{when} the fairness intervention is introduced during the decision making pipeline; 
\textit{Pre-processing} techniques try transforming the training data before feeding it into the ML algorithm.
\textit{In-processing} techniques try to adjust the ML algorithm to include fairness considerations during training.
In contrast, \textit{post-processing} is performed after training, which means that it is applicable even if the training data or the learning algorithm cannot be modified (see~\cite{caton2020fairness} for a more detailed comparison of the techniques).
}
In all cases, it is assumed that a prediction model predicts an unknown but decision-relevant attribute $Y$, that the decision is a function of the predicted value, and that a specified fairness constraint (FC) must be satisfied.
One popular technique to ensure such a FC is to \textit{post-process} the model's output to derive optimal decision rules, representing a function that takes the predictions (and possibly other attributes) as an input and outputs a decision for each individual~\cite{hardt2016equality,10.1145/3097983.3098095,baumann2022sufficiency}.
By formulating algorithmic fairness as a constrained optimization problem, such optimal fair decision rules can be obtained.

In this paper, we present three main contributions:
First, drawing on ML and moral philosophy literature, we introduce a step-by-step procedure for the moral assessment and implementation of group fairness in prediction-based decision making systems.
We hope that this guides practitioners in their goal to implement fair data-based decision making systems while balancing the tradeoffs inherent to this endeavor.
Second, we extend the analysis of the FEC principle introduced by~\cite{Loi2019}.
In particular, we provide moral definitions for several fairness criteria that have not been considered by the authors, making them applicable for the choice of any of the well-known group fairness criteria.
Third, we introduce an extended version of the FEC principle to account for morally irrelevant values of the justifier (which represents an attribute that justifies unequal distributions of a benefit across groups).
We show that this results in relaxations of the common group fairness criteria.

\section{Related work}

Numerous statistical fairness criteria have been proposed to assess the fairness of decision making systems that rely on supervised ML to make consequential decisions~\cite{verma2018fairness,castelnovo2021zoo}.
One line of work strives for fairness across different groups, which resulted in many different so-called \textit{group fairness criteria}.
To choose a morally appropriate criterion, an ethical fairness principle is needed to justify a specific definition of fairness for any given context.
The algorithmic fairness literature still lacks a fairness principle, which is general enough to grasp all relevant group fairness criteria.

Optimal fair decision rules can be derived by formulating algorithmic fairness as a constrained optimization problem~\cite{hardt2016equality,10.1145/3097983.3098095,baumann2022sufficiency}.
Thereby, the performance representing the overarching goal (measured with a utility function specifying the desirability of an outcome from the perspective of the decision maker) is maximized while still satisfying a FC, which is measured with a specific fairness metric~\cite{mitchell2021algorithmic}.
This results in a tradeoff between the performance and the fairness of such systems~\cite{feldman2015disparateimpact,lipton2018does,friedler2019comparative,Kearns2019EthicalAlgorithm,DBLP:journals/corr/KleinbergMR16}.
Relaxing the FC, requiring a partial fulfillment of a fairness criterion, balances this tradeoff.
The most famous example of relaxing a group fairness criteria is arguably the \textit{four-fifths-rule} advocated by the US Equal Employment Opportunity Commission, which requires the decision rate of the worst-off group to be at least four-fifths of the best-off group's decision rate (representing a relaxation of independence)~\cite{US1979four-fifths-rule-80-percent}.
The \textit{four-fifths-rule} can be generalized by requiring that the ratio between the different groups' percentages of individuals assigned a positive decision is above a certain percentage (called \textit{p\%-rule} by~\cite{zafar2017fairness-constraints}).
Optimal decision rules (i.e., those that maximize the decision maker utility while satisfying a given FC to some degree) form the Pareto frontier, which represents this tradeoff~\cite{kearns2019subgroup,kearns2018gerrymandering}.

Many contributions in the fair ML literature focus on one specific problem, such as specifying a morally appropriate definition of fairness for a particular context~\cite{Binns2018lessons}, translating philosophical notions of fairness to a mathematical representation taking the form of a fairness criterion~\cite{Loi2019}, or implementing a specific definition of fairness~\cite{hardt2016equality,10.1145/3097983.3098095,baumann2022sufficiency}.
However, in practice, all aspects must be considered to morally assess and implement group fairness in prediction-based decision making systems.
To our knowledge, a general procedure combining all of these steps to develop fair prediction-based decision making systems has not yet been proposed.

\section{A step-by-step procedure for fair prediction-based decision making}

We consider prediction-based decision making systems taking decisions $D=\{0,1\}$ for individuals of a population, based on a prediction of a single binary attribute $Y=\{0,1\}$ of the individuals, whose value is decision-relevant but unknown at the time of decision. 
Furthermore, we assume that a prediction model exists that outputs the probability $p$ denoting an individual's probability of $Y=1$ and that there is a decision rule that maps this probability into a binary decision $D$.
We assume that the decision rule is designed to maximize the expected utility of the decision maker, 
which is a function of $D$ and $Y$.
As the latter is not known at the time of decision making, the \emph{expected value} of the utility determines the decision, which is in line with the principle known as \textit{rational decision making}~\cite{VonNeumann2007}.

Fairness is a social desideratum that is not directly related to the decision maker's goals (utility) and, therefore, not automatically incorporated into the decision making process.
If fairness is to be considered in the decision making, there is a need to operationalize the concept of fairness.
This includes that fairness has to become measurable and that is implemented into the decision making.

In this paper, we build on a principle of justice proposed explicitly for prediction-based decision making, called \textit{Fair Equality of Chances} (FEC)~\cite{Loi2019}.
This is a general framework that integrates the group fairness criteria separation and sufficiency by giving moral definitions for each of them.
FEC defines fairness generally through three fundamental moral categories: the benefit $U_{DS}$, the justifier $J$ (i.e., the moral claim to $U_{DS}$), and the group membership $G$%
\footnote{
In contrast to the other two categories, the group membership $G$ is commonly used in the fair ML literature -- where it is usually called \textit{sensitive} or \textit{protected attribute} (or simply $A$). 
}.
The goal is to achieve a fair distribution of this benefit $U_{DS}$ across the different groups $G$ by comparing all subgroups of individuals who are equal in their values for $J$.
Formally, FEC is defined as follows:
\begin{definition}[Fair Equality of Chances (FEC) by~\cite{Loi2019}]
An imperfect prediction-based decision rule satisfies FEC if and only if individuals equal in their values for $J$ have the same expectations of $U_{DS}$, irrespective of their $G$ values.
Or equivalently: $\forall g,g' \in G, \forall j \in J, E(U_{DS}|G=g,J=j)=E(U_{DS}|G=g',J=j)$.
\label{def:FEC}
\end{definition}

\begin{figure}[t]
\centerline{\includegraphics[width=0.49\textwidth]{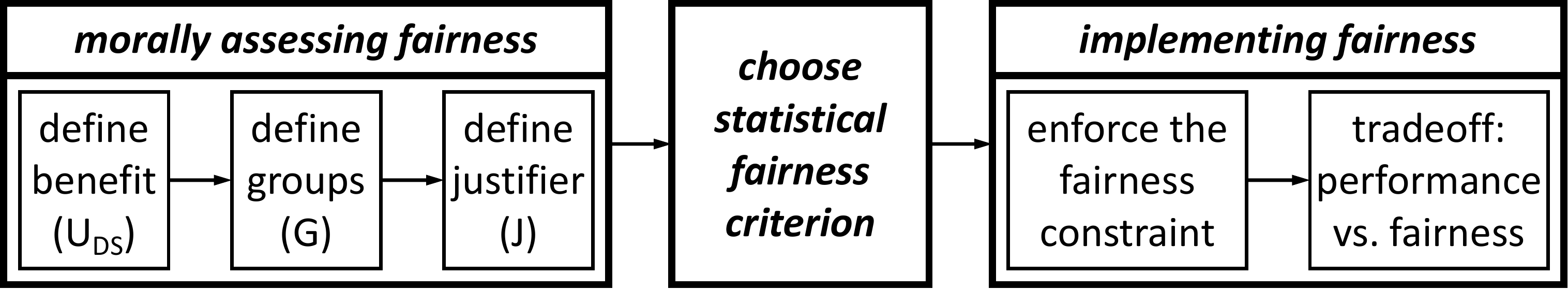}}
\caption{A step-by-step procedure for fair prediction-based decision making}
\label{fig:step-by-step-procedure}
\end{figure}

In this work, we provide a step-by-step procedure to ensure the fairness of prediction-based decision making systems.
This procedure enables data scientists to implement fairness by choosing a decision rule that is optimal regarding the underlying business goal while at the same time satisfying a definition of fairness, which is well-specified based on a thorough ethical analysis.
Before being able to implement fairness into a decision making system, one must define how fairness can be measured, which, in turn, follows from a moral assessment of the system and its context.
Thus, the proposed procedure compromises three components: moral assessment of fairness, choice of a statistical fairness criterion, and implementation of fairness.
The components further consist of consecutive steps, as visualized in Fig.~\ref{fig:step-by-step-procedure}.
In the following sections, we will elaborate on each of these steps in detail.

\section{morally assessing fairness}

This section introduces a step-by-step procedure to morally assess the fairness of a prediction-based decision system, yielding a morally appropriate definition of fairness in the given context.
In practice, this procedure consists of defining the moral categories suggested by the (extended) FEC principle, which can then be mapped to a statistical group fairness criterion in the subsequent step.
It is this mapping that ultimately allows assessing the fairness of the decision making system.
It can be applied to arbitrary application contexts.

\subsection{Define Benefit}

Group fairness always means the fair distribution of ``something'' among groups as a result of applying a decision rule.
In the FEC principle, this ``something'' is called \emph{benefit} for simplicity, but this also includes harm (negative benefit) or opportunity (the possibility of a benefit).
So, we need to define this benefit, and a natural way to do this is as the \emph{utility from the perspective of the affected individuals} $U_{DS}$, where $DS$ stands for decision subjects.%
\footnote{
This is equivalent to what~\cite{heidari_moral_2019} call the \textit{difference between an individual's actual and effort-based utility}.
The FEC principle is based on the \textit{actual} benefit.
However, as the \textit{effort-based} utility is constant for individuals who are equal in their value for $J$, equal expected \textit{actual} benefits is equivalent to equal expected differences between \textit{actual} and \textit{effort-based} utilities.
}
Similar to the decision maker's utility, this benefit may depend on both the decision $D$ and the value of $Y$.
For imperfect decision rules, its expected value is to be considered.



\subsection{Define Groups}

Another part of the moral analysis is to define the groups considered for the comparison, e.g., men vs. women or groups of different ethnicities.
The groups are denoted by $G$, and it is assumed that they are morally arbitrary, meaning that there is no moral justification for receiving unequal benefits just because of being part of this group.
In most application contexts, these are social groups that are defined by non-discrimination principles or legislation. 
From a formal perspective, the attribute indicating an individual's group membership must be observable.%
\footnote{
Despite being a straightforward approach, omitting sensitive attributes from the training data (also known as \textit{fairness through unawareness} or \textit{blinding}) does not imply fairness due to redundant encodings of the sensitive attribute (which is likely for large datasets)~\cite{pedreschi2008discrimination-aware}.
}
However, it is irrelevant whether or not the attribute defining the group membership has been used as an input to train the prediction model.
What matters are the outcomes the decision rule produces w.r.t. these groups.

\subsection{Define Justifier}

The general idea of the FEC principle is that advantages (or disadvantages) associated with a decision making system should be distributed fairly across groups.
However, a fair distribution is not necessarily one that grants each individual the same expected value of $U_{DS}$.
There might be moral reasons implying that some individuals deserve more benefit than others.
Accordingly, one must specify what, if anything, \emph{morally} justifies unequal benefits for decision subjects by the system.
In other words, it must be assessed whether there is an attribute that \textit{justifies} an inequality of benefits for some of the decision subjects.
If such an attribute exists, we shall call it the justifier $J$, which stands for something that makes inequalities in the expected benefit \textit{just}.
Equality of the expected benefit for different groups $G$ is only required for subgroups of individuals who are equal in their value of $J$.%
\footnote{
Suppose a bank must decide which applicants to grant a loan, i.e., $D=1$ denotes a positive decision (granting a loan to an applicant), and $D=0$ denotes a negative decision.
However, the applicant's ``true type'' $Y=\{0,1\}$ is not known at the time of decision (individuals of type $Y=1$ repay a granted loan, and those of type $Y=0$ would default).
Further, suppose the bank uses a prediction model to estimate each applicant's repayment probability.
In this case, one could argue that an individual's moral claim to receive a benefit depends on its type; thus, $J=Y$.
This choice of the justifier $J$ ensures that only individuals who are equal w.r.t. $Y$ are compared across groups when assessing the fairness of the decision making system.
}

The notion of a justifier $J$ is \textit{formally} equivalent to what~\cite{heidari_moral_2019} call the \textit{accountability features}.
However, \textit{morally} speaking, the $J$ used in the FEC principle is a generalization of what is called \textit{accountability} in~\cite{heidari_moral_2019} because it is also compatible with something other than \textit{accountability}, for example, \textit{need}, \textit{merit}, \textit{responsibility}, or \textit{desert}~\cite{wiggins1998needs,roemer2002equality-of-opportunity,feldman2016distributive-justice}.

\section{Choose statistical fairness Criterion}
\label{ssec:Choose_Metric}

The choice between different group fairness criteria can be mapped to specific assumptions regarding the outcome that the decision subjects morally deserve.
Depending on how the moral categories $U_{DS}$ and $J$ have been defined for a specific context, a different group fairness criterion is appropriate.
In the following, we explain how the FEC principle can be utilized to choose a morally appropriate group fairness criterion, and we extend the FEC principle so that its application encompasses all well-known group fairness criteria.
Table~\ref{table:group-fairness-metrics} maps specific choices for $U_{DS}$ and $J$ to well-known group fairness criteria for an example of two groups $G=m$ and $G=f$.%
\footnote{
Note that the distributed benefit $U_{DS}$ must either be related to the decision $D$ or the decision-relevant attribute $Y$ -- otherwise, none of the considered group fairness criteria are morally appropriate. The justifier $J$ can be specified based on any observable attribute, as long as $J$, $G$, and $U_{DS}$ are unequal.
}
\begin{table*}[t]
\centering
\caption{Choosing the most appropriate group fairness criteria for a given context according to the E-FEC principle w.r.t. $G=\{m,f\}$}
\label{table:group-fairness-metrics}

\begin{tabular}{lllll}
\textbf{Group fairness criterion} & \textbf{$U_{DS}$}    & \textbf{$J$}         & \textbf{$j^*$}                               & \textbf{Mathematical representation}          \\ 
\hline
Independence$^\mathrm{a}$      & \multirow{2}{*}{$D$} & --                   & --                                           & $P(D=1|G=m)=P(D=1|G=f)$                       \\ 
\cdashline{1-1}\cdashline{3-5}
Conditional statistical parity &                      & $L$                  & \begin{tabular}[c]{@{}l@{}}--\\\end{tabular} & $P(D=1|L=l,G=m)=P(D=1|L=l,G=f)$               \\ 
\hdashline
Separation$^\mathrm{b}$        & \multirow{3}{*}{$D$} & \multirow{3}{*}{$Y$} & --                                           & $P(D=1|Y=i,G=m)=P(D=1|Y=i,G=f), i \in \{0,1\}$\\ 
\cdashline{1-1}\cdashline{4-5}
TPR parity$^\mathrm{c}$        &                      &                      & $\{1\}$                                      & $P(D=1|Y=1,G=m)=P(D=1|Y=1,G=f)$               \\ 
\cdashline{1-1}\cdashline{4-5}
FPR parity$^\mathrm{d}$        &                      &                      & $\{0\}$                                      & $P(D=1|Y=0,G=m)=P(D=1|Y=0,G=f)$               \\ 
\hdashline
Sufficiency$^\mathrm{e}$       & \multirow{3}{*}{$Y$} & \multirow{3}{*}{$D$} & --                                           & $P(Y=1|D=j,G=m)=P(Y=1|D=j,G=f), j \in \{0,1\}$\\ 
\cdashline{1-1}\cdashline{4-5}
PPV parity$^\mathrm{f}$        &                      &                      & $\{1\}$                                      & $P(Y=1|D=1,G=m)=P(Y=1|D=1,G=f)$               \\ 
\cdashline{1-1}\cdashline{4-5}
FOR parity                     &                      &                      & $\{0\}$                                      & $P(Y=1|D=0,G=m)=P(Y=1|D=0,G=f)$               \\ 
\hline
\multicolumn{5}{l}{Equivalent criteria: $^{\mathrm{a}}$statistical parity, demographic parity, group fairness, $^{\mathrm{b}}$equalized odds, disparate mistreatment, $^{\mathrm{c}}$equality of opportunity, false}           \\
\multicolumn{5}{l}{negative error rate balance, $^{\mathrm{d}}$false positive error rate balance, predictive equality, $^{\mathrm{e}}$conditional use accuracy equality, $^{\mathrm{f}}$predictive parity, outcome test}                                                                                                     

\end{tabular}
\end{table*}

\subsection{FEC and the Choice of a Group Fairness Criterion}
\label{sssec:FEC-and-the-Choice-of-a-Group-Fairness-Metrics}

The FEC principle as introduced by~\cite{Loi2019} is restricted to the two group fairness criteria separation and sufficiency.
The conditions of the FEC principle are equivalent to those of \emph{separation} if $(U_{DS}=D) \land (J=Y)$.
Separation requires that, among all those individuals who are equal in $Y$, an equal share of individuals is assigned a positive decision across groups.
For this criterion to be appropriate, receiving $D=0$ or $D=1$ must produce a benefit for a decision subject, and $J$ must depend on the decision-relevant attribute $Y$, i.e., everyone with the same value for $Y$ equally deserves to receive $U_{DS}$.

The conditions of the FEC principle are equivalent to those of \emph{sufficiency} if $(U_{DS}=Y) \land (J=D)$.
Sufficiency requires that, among all those individuals that are assigned the same decision $D$, an equal share of individuals belongs to the positive class ($Y=1$) across groups.
Sufficiency is a morally appropriate notion of fairness if the decision-relevant attribute $Y$ produces a benefit for the decision subjects and if everyone who is assigned the same decision $D$ equally deserves to receive this benefit $U_{DS}$.

We now extend the findings of~\cite{Loi2019} and show that the FEC principle can also be used for justification of other group fairness criteria (independence and conditional statistical parity), depending on the specification of the moral categories $U_{DS}$ and $J$.%
\footnote{
In very specific situations where the expected utility of the decision subjects is constant for a subgroup formed by a value $J=j$, mapping a morally appropriate definition of fairness to a statistical group fairness criterion results in a relaxation of separation (or sufficiency).
For example, suppose separation is a morally appropriate definition of fairness and that $E(U_{DS}(Y=0, D=0)) = E(U_{DS}(Y=0, D=1))$.
In this case, individuals of type $Y=0$ are indifferent about which decision they are assigned, i.e., FEC is always satisfied for them, meaning that FPR parity is superfluous for fairness and that TPR parity should be applied.
}
\emph{Independence} requires an equal share of individuals to be assigned a positive decision ($D=1$) across groups $G$.
This group fairness criterion is morally appropriate if receiving a positive or a negative decision is what produces a benefit $U_{DS}$ for an affected individual ($U_{DS}=D$) and if everyone equally deserves to receive or not to receive this benefit.
In this case, no justifier exists, i.e., $J=\text{none}$.
\emph{Conditional statistical parity} restricts the parity requirement to a specific subpopulation defined by a set of observable attributes $L$ (called ``legitimate factors'' by~\cite{10.1145/3097983.3098095}).
Hence, this group fairness criterion only requires parity of expected decisions among those subgroups of $G$ who are equal in their values for $L$.

\subsection{Extending the FEC Principle}

The FEC principle requires equality of the expected $U_{DS}$ for all values of $J$ and, thus, its application is limited to situations in which all values $j \in J$ are morally relevant.
We here extend the FEC principle to also account for morally irrelevant values of the justifier $J$:
\begin{definition}[Extended Fair Equality of Chances (E-FEC)]
An imperfect prediction-based decision rule satisfies E-FEC if and only if individuals equal in their morally relevant values for $j^* \in J$ have the same expectations of $U_{DS}$, irrespective of their $G$ values.
Or equivalently: $\forall g,g' \in G, \forall j^* \in J , E(U_{DS}|G=g,J=j^*)=E(U_{DS}|G=g',J=j^*)$.
\label{def:FEC_extended}
\end{definition}
The E-FEC principle is to be applied whenever some values $j'$ of $J$ are deemed to be morally irrelevant with respect to fairness, i.e., inequality of chances for those values $j'$ are not considered a fairness issue. 
In such cases, the FEC equality principle is restricted to a specific value $j^*$ of $J$, e.g., only applied to $J=1$ (but not to $J=0$), which leads to a relaxation of the fairness criterion.%
\footnote{
True positive rate (TPR) parity and false positive rate (FPR) parity are relaxations of separation: TPR parity requires equal decision rates for individuals of type $Y=1$ across $G$, and FPR parity requires equal decision rates for individuals of type $Y=0$ across $G$.
Positive predictive value (PPV) parity and false omission rate (FOR) parity are relaxations of sufficiency:
PPV parity requires individuals who are assigned $D=1$ to be equally likely to belong to the positive class $Y=1$ across $G$.
FOR parity requires individuals who are assigned $D=0$ to be equally likely to belong to $Y=1$ across $G$.
}
This can best be illustrated with an example for separation: We consider a bank granting loans to individuals based on a model that predicts the repayment probability of these applicants.
Let $D=1$ denote a positive decision, meaning that the loan is granted ($D=0$ represents \textit{not} granting a loan), and $Y=1$ means the loan would be repaid if granted ($Y=0$ represents a default).
We assume that $Y$ is considered a morally legitimate justifier ($J=Y$), i.e., individuals repaying their loan morally deserve a positive decision, which is not the case for individuals who would not repay.
This leads to separation. 
In such a situation, the requirement of a fair distribution of the benefit of receiving a bank loan might be restricted to the ``deserving'' individuals ($Y=1)$, and FEC boils down to TPR parity as a relaxation of separation.
The fact that an imperfect decision rule also creates false positives would be considered irrelevant w.r.t. fairness.%
\footnote{One might, of course, argue that granting a loan to someone who cannot repay generates a harm to this individual, and that fairness also includes an equal distribution of this harm. In this case, a relaxation to TPR would not be appropriate. This shows again that the choice of the fairness criteria depends on the moral assessment of what is distributed and for which individuals equality of this distribution should be ensured. Different outcomes of this moral assessment lead to different statistical fairness criteria.}

\section{implementing fairness}

\subsection{Enforce the Fairness Constraint}
\label{ssec:Enforce-the-Fairness-Constraint}

We consider \textit{post-processing} solutions for prediction-based decision making systems to maximize the decision maker utility ($U_{DM}(D)$) while satisfying a fairness constraint (FC).
The optimal decision rule for unconstrained decision makers does not consider fairness and thus may or may not produce unfairness.
To ensure algorithmic fairness, the decision strategy must be adopted so that a specific FC is satisfied, as formulated in the following constrained optimization problem:
\begin{equation}
\small
\argmax_d E(U_{DM}) \text{ subject to FC},
\label{equation:constrained_optimization_problem}
\end{equation}
where FC represents the group fairness criterion that has been chosen in the previous step of the procedure (see Section~\ref{ssec:Choose_Metric}).

The optimal decision rule depends on the chosen definition of fairness.
\cite{hardt2016equality} and~\cite{10.1145/3097983.3098095} provide solutions for (conditional) statistical parity and separation (including its relaxations).
They prove that the optimal decision rule $d^*$ satisfying one of these FC always takes the form of group-specific thresholds:
\begin{equation}
\small
d^*=\begin{cases}
    1 & \text{$p \geq \tau_g$} \\
    0 & \text{otherwise}
\end{cases}
\label{equation:optimal_decision_rules_under_FC_hardt_corbett-davies}
\end{equation}
where $\tau_g \in [0,1]$ denote constants that differ depending on the group membership $g \in G$.%
\footnote{
When choosing conditional statistical parity as the FC, these constants also depend on the ``legitimate'' attributes $L$ (see Table~\ref{table:group-fairness-metrics})~\cite{10.1145/3097983.3098095}.
}
Optimal decision rules satisfying separation may require randomization because TPR parity does not imply FPR parity and vice versa (for details see~\cite{hardt2016equality}). 

\cite{baumann2022sufficiency} provide solutions for sufficiency and its relaxations (PPV parity and FOR parity) and prove that the optimal decision rule $d^*$ satisfying one of these FC always takes the form of group-specific lower- or upper-bound thresholds:
\begin{equation}
\small
d^*=\begin{cases}
    1 & \text{$p \in [\tau_g^l, \tau_g^u]$} \\
    0 & \text{otherwise}
\end{cases}
\label{equation:optimal_decision_rules_under_FC_sufficiency}
\end{equation}
where $\tau_g^l$ and $\tau_g^u$ denote constants representing the lower- and the upper-bound that differ depending on the group membership $g \in G$ and either $0 < \tau_g^l < \tau_g^u = 1$ (for a lower-bound threshold rule) or $0 = \tau_g^l < \tau_g^u < 1$ (for an upper-bound threshold rule) holds. 
As for separation, enforcing sufficiency may require randomization for one group to balance the PPV and FOR across groups (for details see~\cite{baumann2022sufficiency}).


\subsection{Tradeoff: Performance vs. Fairness}

There is an inherent tradeoff between the performance and the fairness of a prediction-based decision system.
This tradeoff can be balanced by considering a partial fulfillment of the chosen FC, for example, by requiring minimum ratios instead of equality.%
\footnote{
Note that instead of ratios, one could also consider differences, e.g., constraining the maximum allowed difference between the groups' FPR.
}
For example, for the FC FPR parity, this can be written as:
\begin{equation}
\small
\text{min}\!\left(\!\frac{P(D=1|Y=0,G=f)}{P(D=1|Y=0,G=m)}, \frac{P(D=1|Y=0,G=m)}{P(D=1|Y=0,G=f)}\!\right) \geq \gamma,
\label{equation:partial_fulfillment_of_fairness}
\end{equation}
where $\gamma \in [0,1]$ and $\gamma = 0$ represent an unconstrained optimization problem whereas $\gamma = 1$ represents parity of TPR across groups.
All group fairness criteria defined in Table~\ref{table:group-fairness-metrics} can be relaxed to take a similar form as Eq.~\ref{equation:partial_fulfillment_of_fairness}.
Deriving optimal decision rules for different values of $\gamma$ makes the tradeoff between performance and fairness tangible.

Some legislation relaxes fairness requirements from strict equality to approximate equality, e.g., by applying the \textit{p\%-rule}~\cite{zafar2017fairness-constraints}. This corresponds to a specific value for $\gamma$.


\section{Application}

In this section, we showcase how the proposed step-by-step procedure can be applied to a prediction-based decision making system, using the example of \textit{jail-or-release} decisions taken by judges.
We trained a logistic regression on the ProPublica recidivism dataset%
\footnote{
We used the pre-processed dataset provided by~\cite{friedler2019comparative}, which can be accessed here: \url{https://github.com/algofairness/fairness-comparison/tree/master/fairness/data/preprocessed} (see file ``\textit{propublica-recidivism\_numerical.csv}'').
}, including data from the COMPAS tool collected by~\cite{angwin2016machine}, to predict probabilistic recidivism risk scores.
Here, $Y=1$ indicates that a released individual would reoffend, and $Y=0$ otherwise.
Based on these predictions, judges must decide whether a defendant is granted bail or not (i.e., the defendant is given the decision $D=0$ if released and $D=1$ if detained).
We implemented a \textit{post-processing} solution to derive optimal fair decision rules based on a training set (2/3) and evaluated the resulting fairness and performance on the test set (1/3).%
\footnote{
To obtain more reliable results, we split the data into a training and test set 10 times and here present the average values for fairness and performance.
}

The first step to morally assess fairness is to define the benefit that is distributed.
In this context, the benefit manifests itself in the decision, i.e., $U_{DS}=(1-D)$: Individuals who are released ($D=0$) are better off than those detained ($D=1$).
In this example, we assume that the attribute $G$ is race, 
defining \textit{Caucasian} subjects by $G=c$, and \textit{African American} subjects by $G=a$. 
Next, we need to assess justifiers: We argue that it is morally justifiable that $J=Y$ because detainees should only be released if they do not recidivate, otherwise they do not deserve to be released.
This moral specification yields separation as FC.
Furthermore, we argue that for an imperfect decision rule, the dominant moral fairness issue is to make sure that the harm of being kept in prison (D=1) if not being a danger to the public safety (Y=0) is distributed fairly among the groups (FPR parity).
In contrast, differences in FNR (persons wrongly released because they reoffend afterward) are morally irrelevant and do not constitute unfairness. 

With this moral assessment, the appropriate group fairness criterion is FPR parity among Caucasian and African American subjects.
This can be interpreted as follows: Defendants of type $Y=0$ deserve the same sentencing outcome $D$, on average, regardless of their group membership $G$.

\begin{figure}[t]
    \centering
    \subfloat[Partial fulfillment of fairness]{\includegraphics[width=0.2345\textwidth]{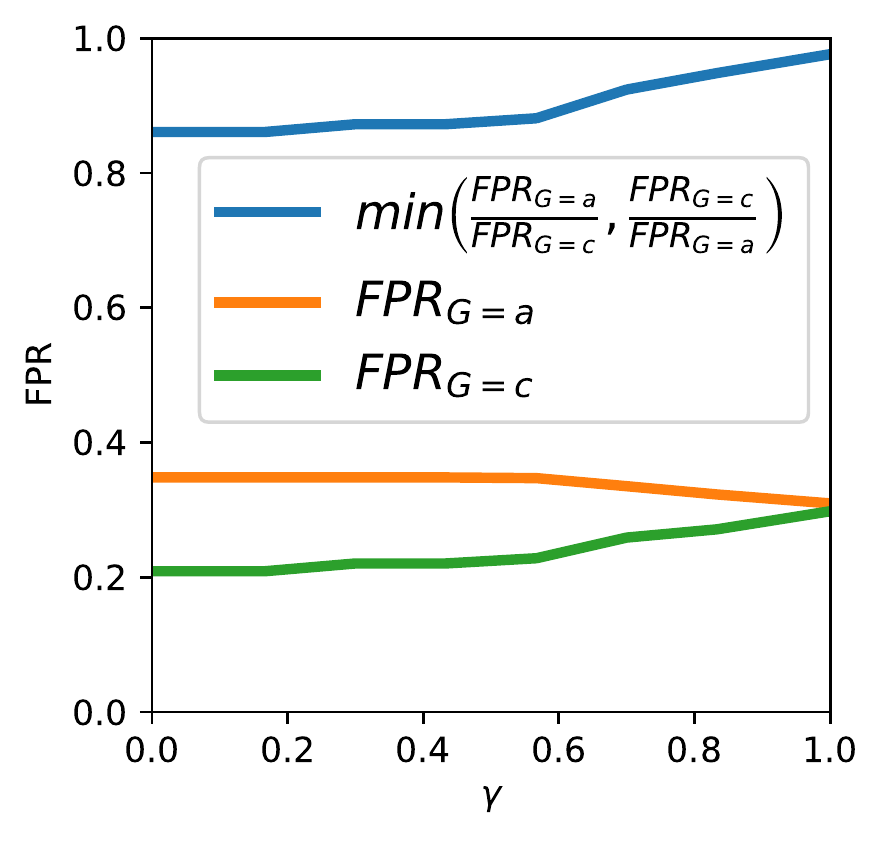}
    \label{fig:gamma_FPR}
    }
    \hfil
    \subfloat[Performance-fairness-tradeoff]{\includegraphics[width=0.24\textwidth]{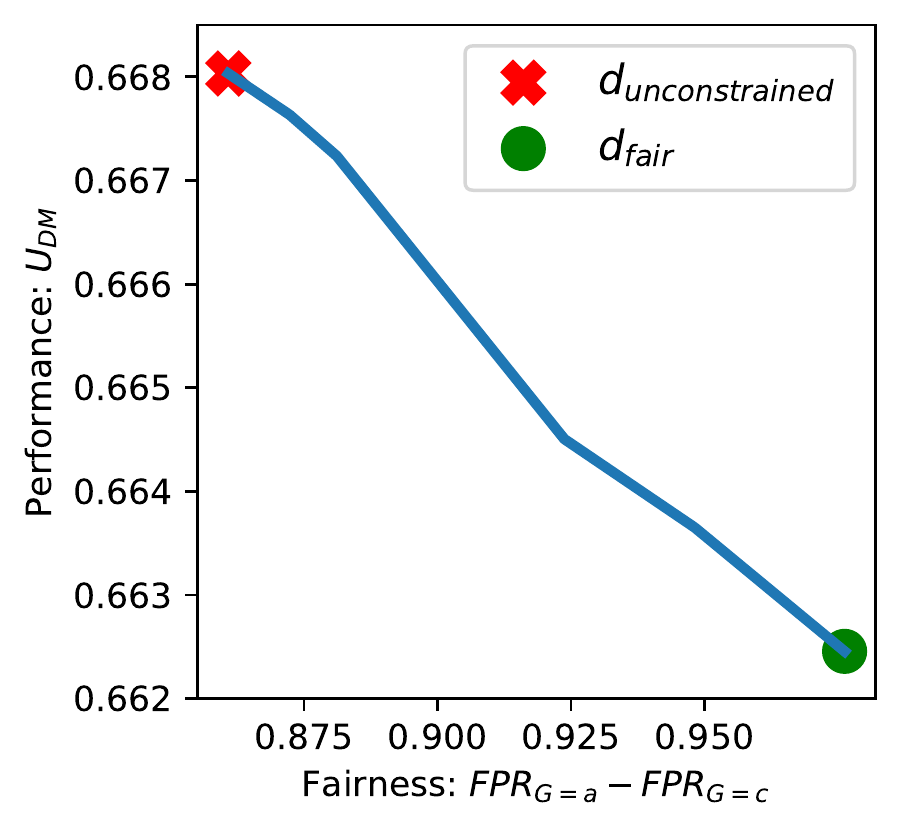}
    \label{fig:pareto_plot}
    }
\caption{Implementing fairness for \textit{jail-or-release} decisions}
\label{fig:COMPAS_solutions}
\end{figure}

To implement fairness in this decision making system, we must solve the utility maximization problem stated in Eq.~\ref{equation:constrained_optimization_problem} subject to the chosen group fairness criterion.
The overarching goal of the decision maker in this context is to maximize public safety, i.e., prevent crimes without detaining innocent people.
For simplicity, we assume that $U_{DM}$ is equivalent to the accuracy of the decisions (assuming that $D=Y$ is the optimal decision), though other utility functions would also be possible.
To balance the tradeoff between performance and fairness, we reformulate the FC according to Eq.~\ref{equation:partial_fulfillment_of_fairness} to allow for partial fulfillment of the constraint instead of requiring full FPR parity.
We then solve the constrained optimization problem s.t. this new FC for different values of $\gamma$.
Depending on different factors (such as existing regulations or stakeholder requirements), one might fully enforce fairness ($\gamma =1$) or also consider a partial fulfillment of the chosen criterion ($\gamma <1$), which results in different TPR, as can be seen in Fig.~\ref{fig:gamma_FPR}.

The performance and fairness resulting from these solutions are visualized in Fig.~\ref{fig:pareto_plot}, representing the cost of fairness for different degrees of fairness.
In addition to this, two specific decision rules are indicated.
The unconstrained optimal decision rule ($d_{unconstrained}$) is a single threshold $\tau=0.5$, which yields $U_{DM}=0.668$, $FPR_{G=a}=0.35$, and $FPR_{G=c}=0.21$.
Enforcing FPR parity yields an optimal decision rule ($d_{fair}$) with group-specific thresholds of $\tau_{a}=0.51$ and $\tau_{c}=0.44$, resulting in a decision maker utility of $U_{DM}=0.662$ and equal FPR $FPR_{G=a}=FPR_{G=c}=0.30$.
Hence, the optimal fair decision rules result in fewer detained African American individuals and more detained Caucasian individuals than the optimal unconstrained decision rule.

\section{Conclusions}

This paper makes several contributions to the field of algorithmic fairness by elaborating on a general fairness principle (FEC), which has been proposed as a basis for morally assessing group fairness in prediction-based decision making systems. The assessment consists of defining what is meant by benefit, which groups we are considering, and which attributes morally justify differences in received benefits.
We showed that the FEC principle naturally leads to the known statistical group fairness criteria statistical parity, conditional statistical parity, separation, or sufficiency, depending on the specification of the moral categories benefit and justifier.
In addition to that, we extended the FEC principle, allowing to account for morally irrelevant values of the justifier, which then leads to specific relaxations of the fairness criteria.
Our work adds to prior work in that we specify the complete process of developing fair prediction-based decision systems, including three essential steps: (a) derivation of a morally appropriate definition of fairness in the given context, (b) mapping of this definition to a fairness criterion, and (c) implementation of this criterion into the decision rule by balancing the tradeoff between performance and fairness.

Our paper thus presents a full framework for mitigating the risk of unfairness associated with the use of prediction-based systems for making consequential decisions.
However, it is important to note that the formalization of the moral assessment that we propose in this paper does not determine the outcome of this assessment, i.e., the fairness criteria. This depends on the context and the moral stances of the involved stakeholders, and, in open societies, people might disagree on how to define the moral categories. However, our approach helps to pose the important fairness-related questions in a structured way, to be explicit in terms of the normative choices that one prefers, and it gives a clear translation of these choices into mathematical fairness criteria.

\section*{Acknowledgment}
We thank Michele Loi and our three anonymous reviewers for their helpful feedback.
This work was supported by Innosuisse -- grant number 44692.1 IP-SBM -- and by the National Research Programme
``Digital Transformation'' (NRP 77) of the Swiss National Science
Foundation (SNSF) -- grant number 187473.
\bibliographystyle{IEEEtran}
\bibliography{IEEEabrv,biblio}

\end{document}